\documentclass[sn-basic]{sn-jnl}% Basic Springer Nature Reference Style/Chemistry Reference Style
\usepackage{graphicx}%
\usepackage{multirow}%
\usepackage{amsmath,amssymb,amsfonts}%
\usepackage{amsthm}%
\usepackage{mathrsfs}%
\usepackage[title]{appendix}%
\usepackage{xcolor}%
\usepackage{textcomp}%
\usepackage{manyfoot}%
\usepackage{booktabs}%
\usepackage{algorithm}%
\usepackage{listings}%
\usepackage{algorithmic}%
\usepackage{float}%
\usepackage{wrapfig}%
\usepackage{soul}%

\def\BibTeX{{\rm B\kern-.05em{\sc i\kern-.025em b}\kern-.08em
    T\kern-.1667em\lower.7ex\hbox{E}\kern-.125emX}}

\begin{document}

\title[Active learning with biased non-response to label requests]{Active learning with biased non-response to label requests}


\author*[1,2]{\fnm{Thomas S.} \sur{Robinson}}\email{t.robinson7@lse.ac.uk}

\author[2]{\fnm{Niek} \sur{Tax}}\email{\{niek,rmudd,idoguy\}@meta.com}

\author[2]{\fnm{Richard} \sur{Mudd}}

\author[2]{\fnm{Ido} \sur{Guy}}

\affil*[1]{\orgdiv{Department of Methodology}, \orgname{London School of Economics},  \orgaddress{\city{London}, \country{UK}}\looseness=-1}

\affil[2]{\orgdiv{Central Applied Science}, \orgname{Meta}, \orgaddress{\city{London, UK}/\country{Tel Aviv, Israel}}}


\abstract{Active learning can improve the efficiency of training prediction models by identifying the most informative new labels to acquire. However, non-response to label requests can impact active learning's effectiveness in real-world contexts. We conceptualise this degradation by considering the type of non-response present in the data, demonstrating that biased non-response is particularly detrimental to model performance. We argue that biased non-response is likely in contexts where the labelling process, by nature, relies on user interactions. To mitigate the impact of biased non-response, we propose a cost-based correction to the sampling strategy--the \textit{Upper Confidence Bound of the Expected Utility (UCB-EU)}--that can, plausibly, be applied to any active learning algorithm. Through experiments, we demonstrate that our method successfully reduces the harm from labelling non-response in many settings. However, we also characterise settings where the non-response bias in the annotations remains detrimental under UCB-EU for specific sampling methods and data generating processes. Finally, we evaluate our method on a real-world dataset from an e-commerce platform. We show that UCB-EU yields substantial performance improvements to conversion models that are trained on clicked impressions. Most generally, this research serves to both better conceptualise the interplay between types of non-response and model improvements via active learning, and to provide a practical, easy-to-implement correction that mitigates model degradation.\vspace{-0.5em}}


\maketitle
\thispagestyle{empty}
\clearpage
\pagenumbering{arabic}

\section{Introduction}
Many real-life machine learning (ML) system deployments contain a human-in-the-loop component where the model is continuously or periodically updated based on new batches of labelled data. These new labels are typically obtained either by a pool of human annotators (e.g., through crowdsourcing) or originate from users' interactions with an application. \emph{Active Learning} (AL)~\citep{settles2009active} is a research field focused on optimizing the efficiency of this labeling, by directing efforts towards instances whose labels are deemed most informative.

Most AL approaches make the critical assumption that every query (i.e., an attempt to acquire the label of an unlabelled example) receives a response. This assumption may not always hold in practice. For example, consider a fraud detection setting~\citep{tax2021machine,carcillo2018streaming} where AL selects which items are to be sent for human review to be investigated for potential fraud or an integrity violation. The responses obtained from the human reviewer are then incorporated into the training data for an automated detection system. In practice, a reviewer may struggle to identify fraud in some instances more so than in others; for instance, emails soliciting payment details may be easier to review than emails containing phishing links. In the harder cases, where the reviewer is unsure, they may avoid returning a label (or deliberately mark it null to indicate that they were unable to reach a conclusion). Beyond highlighting the possibility of non-response, this example also underscores that non-response can be biased, i.e., non-response may be more likely for some unlabelled examples than for others.

Annotations may also arise as an implicit byproduct of users' interactions with an application, rather than from annotators whose main focus is to provide labels (and who are the predominant focus in the AL literature). In implicit labelling cases, AL methods can be used to provide an exploration component to ML applications \citep{elahi2016}. As an example, consider a recommender system that uses AL to present an item to a user that may improve the recommender system's future ability to suggest relevant items \textit{if} we obtain a label for that user-item pair. In such an application, a label may, for example, be obtained only if the user clicks on the recommendation. If, however, the user gets recommended the item but scrolls past it without interacting, this is plausibly an instance of non-response to an AL label request. In such contexts, non-response can be both substantial and also dependent on the user-item pair (i.e., biased). In Section~\ref{sec:taobao} we present a real-life example of this scenario, where labels arise from the actions of users and non-response stems from failures to interact with advertised items.

Some works study \emph{AL with abstention feedback}~\citep{fang2012active,yan_2015,amin2021,NGUYEN2022242}, which accounts for scenarios where the annotator provides no label. However, these studies often overlook the potential consequences of non-response \textit{bias} in the abstention mechanism. While non-response bias is a well-studied problem in statistics, in Section~\ref{sec:sims_nf} we show that its presence in the context of AL with abstention feedback yields new and unique challenges. To illustrate these challenges, consider again the recommender system example. An AL recommender system may query an item for which the model has high uncertainty about the item's relevance to the user. However, the uncertainty for this query may stem from the fact that users rarely interact with this item. Consequently, and in practice, the AL system's attempts to learn about the user's preferences may end up being wasteful as the item likely continues to receive little interaction, which contradicts the goal of AL to maximise model improvement efficiency.

In this paper, we introduce how biased non-response can affect the performance of AL, and demonstrate these mechanisms empirically. We then present an algorithm to adjust AL to account for non-response, demonstrating both experimental and applied contexts in which it improves model performance. We also show specific contexts where model performance is still impacted negatively. In summary, we make three contributions to AL research:
\begin{itemize}
    \item{We conceptualise a mechanism for \emph{how} biased non-response can undermine the supposed benefits of AL and consider important contexts where the non-response probabilities are very high.}
    \item{We propose a simple algorithmic correction for incorporating a model of non-response into AL.}
    \item{We demonstrate how the mechanics of AL can lead to reinforcing negative behaviour due to the unavailability of labels in specific regions of the possible feature space.}
\end{itemize}

\section{Background}
\label{sec:background}
This section introduces the theoretical framework on which our argument rests. We use lowercase letters to denote scalars (e.g., $a$), lowercase bold letters to denote vectors (e.g., $\mathbf{a}$), and uppercase bold letters to denote matrices (e.g.,  $\mathbf{A}$).

We consider the case where a researcher faces a common modelling problem: given a vector of features $\mathbf{x}$, what is the corresponding label $y$? To answer this question, typically the researcher fits a \textit{target} model $\mathcal{M}$ on a training set $\mathbf{D}^\text{Train} = \{\mathbf{X},\textbf{y}\}$, and uses the resulting model to predict labels for new observations $\mathbf{X'}$: 
$$
\hat{\mathbf{y}'}=\mathcal{M}(\mathbf{X'} \mid \mathbf{D}^\text{Train}).
$$

AL extends this logic by making iterative attempts to acquire new labelled examples to improve model performance efficiently. Let $\mathbf{X} \subseteq \mathbb{R}^d$ be a dataset of all instances that can potentially be labelled. At each point in time $t$, we have a training set $\mathbf{D}^\text{Train} = \{(\mathbf{x}_i, y_i)\}_{i=1}^{|T|}$, where each $x_i \in \mathbf{X}$ and $y_i \in \{0, 1\}$. The subset of $\mathbf{X}$ that is unlabelled at time $t$ is the ``pool" ($\mathbf{X}^\text{Pool}$). Given this context, AL first identifies which unlabelled example ($\mathbf{x}_t \in \mathbf{X}^\text{Pool}$) to query:
$$
\mathbf{x}_t \sim \phi(\textbf{X}^\text{Pool},\mathcal{M}_{t-1}),
$$
where $\phi$ is some AL querying strategy given the pool of unlabelled  examples and the previous state of the model. Common criteria include \emph{uncertainty sampling}~\citep{lewis1995} (i.e., identifying the next query include the entropy of the model output), or \emph{Query-by-Committee} (QbC)~\citep{seung1992query,freund1997selective} (i.e., maximising the disagreement between members of an ensemble).

Once the example has been chosen, an annotator (be it a human reviewer or other process) labels this datapoint. In conventional settings, we assume that we receive the observed value of every requested label, $y_t$. More formally, let $\Omega(\cdot)$ be some (unknown) function that determines the probability of receiving a response, such that $p_t = \Omega(\mathbf{x}_t)$ is the response probability for a specific datapoint. Consequently, let $r_t \sim \mathcal{B}(p_t)$, be a draw from the Bernoulli distribution, which indicates whether a response was received for that example. In the conventional setting, therefore, AL assumes that $\Omega(\mathbf{x}_t) = 1, \ \forall{\mathbf{x}_t}$. 

The results of this process are then combined with existing training examples from previous periods:
$$
\mathbf{D}^\text{Train}_t = 
\begin{cases}
    \mathbf{D}^\text{Train}_{t-1} \cup \{\mathbf{x}_t, y_t\}, & \text{if } r_t = 1 \\
    \mathbf{D}^\text{Train}_{t-1}, & \text{otherwise},
\end{cases}
$$
and finally, using the updated training set, a new target model is trained.

\section{Related literature}
\label{sec:related_lit}

Problems associated with missing data have long been documented in the econometrics literature \citep{rubin1976inference}. In practice, data can be missing for a variety of reasons, including imperfect measurement, dataset corruption, and partial responses. In the survey sampling literature, non-response refers to the failure of individuals to reply to a questionnaire or survey \citep{hansen1946problem}, and hence non-response is a particular form of missing data. This longstanding sampling problem has a clear analogue to AL in the form of failed attempts to query a label.

Theoretical work in this area has largely focused on how ``missingness" in the data can \textit{bias} parameter estimates \citep{rubin1976inference,mohan2013graphical,little2019statistical}. In particular, this work has yielded a typology of missingness: data \emph{missing completely at random} (MCAR), where missing values are independent of observed and unobserved features of the data generating process (DGP) including the outcome; data \emph{missing at random} (MAR), where missing values are independent of unobserved features but related to observed features of the DGP; and, data \emph{missing not at random} (MNAR), where missing values are related to unobserved features of the DGP.

Our focus differs from the econometric treatment of missing data because AL is \textit{inherently} biased by the deliberate acquisition of training data \citep{farquhar2021statistical}. By selecting the most informative labels, the data will not follow the underlying population distribution. Thus, our interest is not squarely on the inferential validity of the model, but rather on the relative performance of a model afflicted with missingness compared to the counterfactual context where there is no probability of non-response. Our work is, therefore, more aligned with work on missing data in machine learning contexts, where the goal is to account for missingness to ensure datasets are complete, rather than ascertain unbiased parameter estimates \citep{stekhoven2012missforest}.

A recent corpus of work has considered non-response in AL contexts \citep{fang2012active,yan_2015,amin2021,NGUYEN2022242}. Of these contributions, two provide algorithmic improvements aimed at handling the sorts of non-response described above. \citet{yan_2015} propose repeatedly querying examples with non-response, which may be costly if a null label is highly likely for some informative regions of the feature space. More recent work has sought, therefore, to incorporate the posterior predictive rate of abstention in the objective function \citep{NGUYEN2022242}. The Bayesian aspect of this work is, however, computationally taxing, and in practice requires approximate methods using \textit{maximum a posteriori} estimation and regularised regression models. It is unclear, given these constraints, whether this method can be applied to ensemble-based AL strategies, like QbC, which can be more performant than single-hypothesis methods like uncertainty sampling. In contrast, the method proposed in this paper simply requires weighting the sampling ``score" from any strategy.

Moreover, these works do not formalise directly how differences in the \textit{types} of missingness that determine non-response impact the performance of AL. Work in this area has noted ``knowledge blind-spots'" \citep{fang2012active} and differences when non-response is close and far from the decision boundary \citep{NGUYEN2022242}, but not compared more general, theoretically derived mechanisms of missingness. Finally, we consider AL contexts where a non-response model can be trained separately, and often in advance, which may be especially beneficial in production systems and/or where the rate of non-response is particularly high.

One other proximal research area within AL research focuses on noisy or imperfect labels \citep{NIPS2016_dd77279f}. Here, like in the case of non-response, model performance may be affected by differences between the returned and the true label. However, unlike noisy labels, we know when we get non-response (a null value is returned), whereas we often do not know which labels are noisy, leading to systematic differences in how these complications are handled. Work explicitly on noisy labels has focused on identifying instances where we are unsure about the label value, and re-labelling these points \citep{sheng2008get,zhao2011incremental,lin2016re,nguyen2020clara}.

Finally, there are conceptual similarities between AL with non-response and research on multi-armed bandits~\citep{lattimore2020bandit}, i.e., a class of algorithms that study the setting where an agent iteratively takes an action that results in an observed reward, where the goal is to maximise the total reward over a time window. Bandit algorithms face the choice of \textit{exploring} actions where currently little is known, and \textit{exploiting} existing knowledge. There is a strong connection between the AL problem and \emph{best arm identification} (BAI) ~\citep{audibert2010best}, a bandit context where one is solely concerned with maximising the knowledge about an arm's (potentially context-dependent) reward distribution. One notable difference is that BAI is concerned with exploration for the sake of learning the reward distribution over actions depending on context while AL is concerned with exploration for the sake of minimising expected future prediction errors. Moreover, the \emph{partial monitoring} literature~\citep{bartok2014partial} considers bandit settings where for some actions the reward is never observed. This setting has similarities to an active learning context with zero response probability in certain regions of covariate space. 

\section{The impact of (biased) non-response}
\label{sec:theory}

Labelling takes different forms, which can lead to different non-response mechanisms. First, consider the conventional context where an unlabelled example is sent to a human annotator who explicitly returns a value for that instance. Non-response in these contexts can arise for several reasons. Annotators may have capacity to review a fixed number of labels, smaller than the number requested, and thus some queries remain unreviewed. Or, the annotator may be unsure over the example's label and, rather than guess, abstain or return a null value. Similarly, if there are multiple annotators, a value may not be assigned if there is not majority agreement or consensus.
\begin{figure}[t]
    \centering
    \includegraphics[width = 0.9\textwidth]{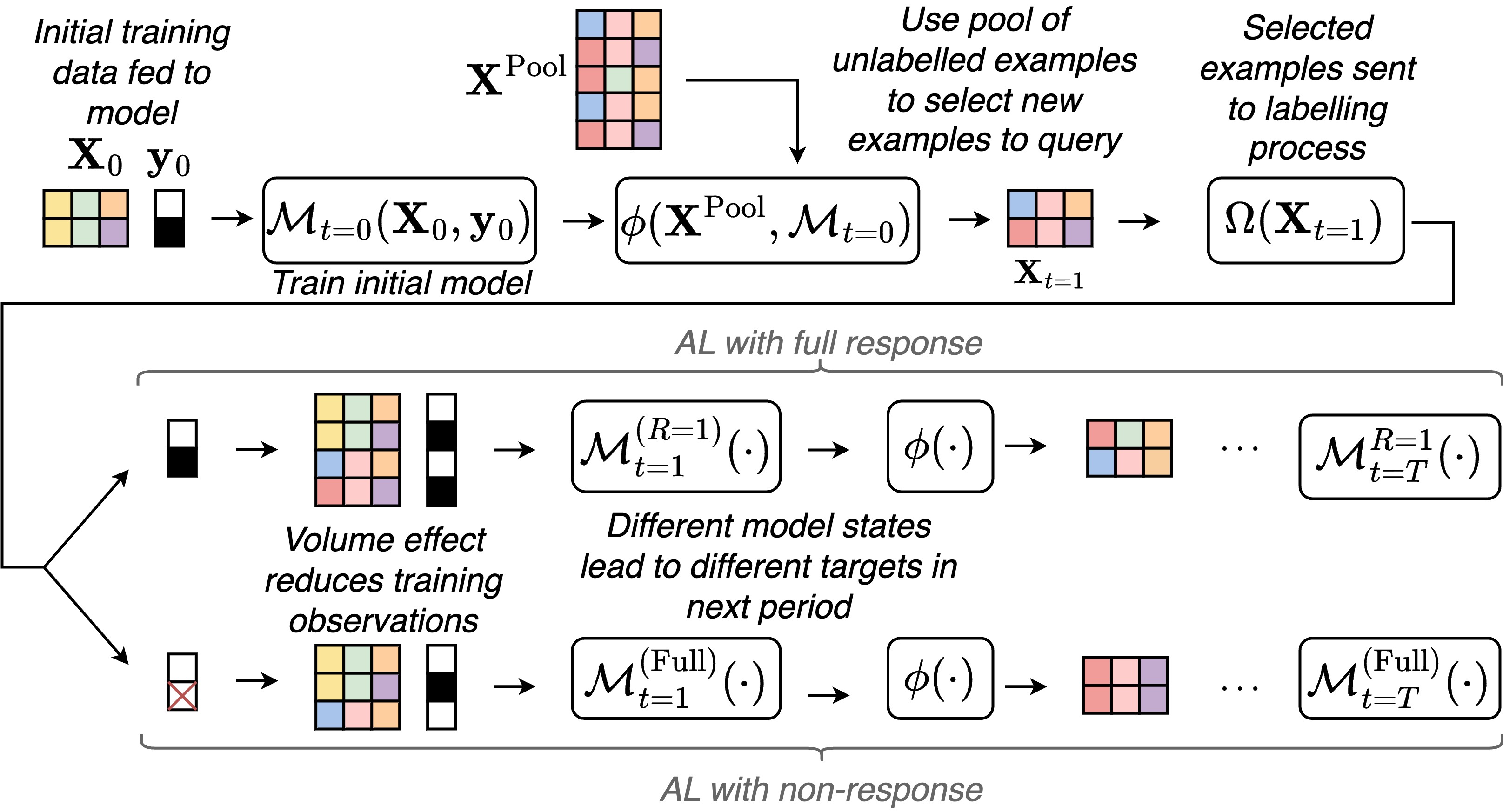}
    \caption{The knock-on consequences of non-response on AL. From the same initial model, non-response leads to volume and imbalance effects in the AL sequence. Here, the result of these effects is the repeated querying of a non-responsive example. Colored blocks refer to data values and the red cross indicates a non-response label. }
    \label{fig:al_pipeline}
\end{figure}
\vspace{-1em}

Not all labelling is explicit, however: annotations can be based on some implicit (user) behaviour. AL is a well-documented method for improving the performance of recommendation systems \citep{elahi2016}. Real-life ranking systems often factorize the target prediction into several components~\citep{ma2018entire}. For example, video streaming services often train two models that separately estimate a \emph{click-through rate} (CTR), $P(\mathit{click}=1)$, and some quantity that depends on the user behavior after the click such as watch time ($E[\textit{watch time} \mid \mathit{click}=1]$). This second model is referred to as the \emph{post-click model}. A final ranking is created by sorting on the result of their multiplication: $E[\textit{watch time}] = E[\textit{watch time} \mid \mathit{click}=1] \times P(\mathit{click}=1)$~\citep{lin2023tree}. 

Similarly, advertisement ranking systems often estimate conversion probabilities of ads by factoring this quantity into a \emph{CTR model} and a \emph{post-click model}, i.e., $P(\mathit{conversion})=P(\mathit{conversion}\mid\mathit{click}=1)P(\mathit{click}=1)$, where the post-click model estimates the conversion probabilities of clicked ads ~\citep{barbieri2016improving,rosales2012post,ma2018entire}. In the ranking systems of these examples we may seek to employ active learning to make the exploration of post-click models estimating $P(\mathit{click}=1)$ and $E[\textit{watch time} \mid \mathit{click}=1]$ more efficient. We discuss a practical example in more depth in Section~\ref{sec:taobao}.

Active learning systems that aim to improve post-click/interaction models are particularly affected by non-response since labels in these settings are contingent upon user interactions such as clicks. In such contexts, the probability of receiving a valid label may be very low as users only interact with a tiny share of the items. Consequently, the costs of AL may be considerably higher and outweigh the benefits of purposefully serving new advertisements or recommendations in order to improve the post-click model.

Beyond the labelling costs of non-response, the presence of missing or null values across both explicit and implicit AL contexts can affect model training, potentially introducing additional sources of bias into the model framework \citep{cortes2018online}. More formally, let $R_{\Omega(\mathbf{x})}$ be a random binary variable that denotes whether the labelling process returns a valid response (1) or non-response (0). Wherever the query is clear from the context, we will refer to this random variable as $R_\Omega$, for simplicity. We assume that a query can be repeated in subsequent rounds if a label is not returned in the current round. As a result, the prediction model becomes conditional not only on the AL-identified training set but also on the success of the labelling process itself, i.e.,
$$
\hat{\mathbf{y'}} = 
\begin{cases}
    \mathcal{M}_t(\mathbf{X'} \mid \mathbf{D}^\text{Train}_t), &
    \text{if } R_\Omega=1, \\
     \mathcal{M}_t(\mathbf{X'} \mid\mathbf{D}^\text{Train}_{t-1}) &
\text{otherwise.}
\end{cases}
$$

Hypothetically, we can quantify the impact that non-response has by considering the differential model performance (using some metric like the ROC AUC score) between (i) a model trained only on the data with responses and (ii) a full-response baseline where all queried examples are labelled. There are two pathways through which non-response may impact this performance:
\begin{enumerate}
    \item Non-response leads to a reduction in the number of training samples for any round $t$.  We call this the ``\textit{volume effect}". Since new examples are only added to the training data when there is a non-null label, the reduction in the volume of examples will reduce the ability of the model to improve, compared to the full-response AL model.
    \item Non-response alters the distribution of training examples, relative to the full-response model, leading to an ``\textit{imbalance effect}". This imbalance affects the performance of the model at time $t$, though at the most general level it is unclear \textit{how} it would do so: bias in the non-response could even (partially) cancel out the inherent selection bias of AL frameworks that we noted earlier. 
\end{enumerate}

Crucially, both the volume and imbalance mechanisms have knock-on effects on model performance in \textit{subsequent} rounds of labelling, since the state of the model in period $t$ determines the selection of new target labels in $t'{>}t$. Where non-response induces poor model performance on certain regions of the feature space, the querying function may seek to address this deficiency by oversampling this area in future rounds. For example, in a video recommender model, we only observe the watch time of a video when the user clicks it. The non-response rate, therefore, will be higher for items with low CTR. Oversampling these regions to try to get better data on watch time, by upranking these videos, simply results in low response rates in future queries.

Figure \ref{fig:al_pipeline} summarises the hypothesised impact that non-response has on the AL framework. In short, a combination of volume and imbalance effects alters the selection of new unlabelled examples, which in turn leads to degradation in model performance. Given the sequential nature of AL training, this loss of performance has the potential to compound over multiple steps of the AL sequence.

\subsection{Types of non-response}

We contend that the mechanism(s) of non-response itself can impact \textit{how} and the extent to which these two non-response dynamics -- the training and selection effects -- impact model performance. Here we focus on two types of non-response, applying intuitions from the wider inferential literature on missing data \citep{rubin1976inference}.

\textbf{Missing completely at random (MCAR)} In the simplest case, the labelling process may be prone to corruption that is random in nature. For example, data loss incurred when streaming data due to unreliable networks or input errors made by human annotators may mean that labels are not always returned. Crucially, in these cases non-response is induced in ways unrelated to the subject of the measurement.

If non-response is MCAR, then:
$$
P(R_{\Omega}=1 \mid \mathbf{X}, \mathbf{y}, \phi) = P(R_\Omega = 1).
$$
That is, the probability of missingness is distributed uniformly across the entire feature space and orthogonal to which unlabelled examples are queried. Therefore, while non-response will reduce the size of the training set, we would not expect any substantial imbalance effect as a result of non-response.

There are many reasons why AL may suffer from MCAR non-response. Consider the human annotator example discussed previously. If any requested labels are not logged by the end of the working day, suppose they are left unlabelled. If the order of labels given to the annotator is random, then this non-response is unrelated to feature values.

\textbf{Missing at random (MAR)} Alternatively, it may be that the non-response mechanism \textit{is} related to $\mathbf{X}$ and/or $\mathbf{y}$. In these instances, the probability of non-response is not uniform across the feature space. Instead, and assuming non-response is explainable by observed features present in $X$, then:
$$
P(R_{\Omega}=1 \mid \mathbf{X}, \mathbf{y}, \phi) = P(R_{\Omega}=1 \mid \mathbf{X}) \neq P(R_{\Omega}=1).
$$

For example, consider the CTR and post-click setting described earlier. Our ``query" might be an advertisement placed in a video carousel, where a label is assigned only if users click through from the advert to the product page. Importantly, features of the user may determine both whether the advert is shown (i.e., the user is queried) \textit{and} whether the user clicks.

As Table \ref{tab:missing_impact} summarises, MAR non-response does not immediately introduce problematic bias into the model. Rather, the effect of missingness on model performance will depend on the interaction between the distribution of \textit{informative} datapoints (at time $t$) and the (unknown) distribution of non-response. 
\begin{table}[b]
    \centering
    \caption{Summary of hypothesised effects of non-response in AL}
    \label{tab:missing_impact}
    \begin{tabular}{cc|cc}
    \toprule
    & & \multicolumn{2}{c}{\textbf{Effect}} \\
    \textbf{Non-response} & \textbf{Cor(Informativeness, $\mathbf{R_\Omega}$)} & \textit{Volume} & \textit{Imbalance} \\
    \midrule
         MCAR & -- & Yes & No \\
         \midrule
         MAR & Negative & \textit{Limited} & No \\
         MAR & Positive & Yes & Yes \\
         \bottomrule
    \end{tabular}
\end{table}
On the one hand, the probability of non-response could be negatively related to the probability of being queried. In other words, some portion of the feature space may be more likely to return missing values, but also have low informativeness from an AL perspective. For example, younger users may be less likely to click through adverts, but given their age, the model is already confident about predicting these conversion outcomes. As a result, AL would favour requesting labels from other portions of the dataset, which are less likely to be affected by non-response. In which case, while there may be some limited volume effect due to a small number of non-responses, we would not expect a substantial biasing imbalance effect.

On the other hand, and perhaps more naturally, if some examples are both highly informative and have high rates of non-response, then \textit{both} a volume and bias effect will impact model performance. This case can be acutely relevant to e-commerce and other content platforms. For example, if one set of items in the advertising model never gets clicked, then the model is likely to be uncertain about their conversion probabilities due to a lack of training examples, but to refine these estimates would exactly require recording clicks. More generally, higher non-response may be precisely why there is uncertainty in these regions, which over the course of sampling iterations leads to reduced informativeness in all parts of the feature space except here. 

As in the MCAR case, non-response reduces the size of the training set, but now this effect is compounded by non-response occurring precisely for those unlabelled examples AL has identified as being most important for model improvement. In the case where the model is able to learn from other parts of the feature space, then over subsequent rounds the relative weight placed on querying this space will increase. Hence, the selection effect may force the model into repeatedly sampling from a region with high non-response rates, which may stall or degrade model improvement.

There is one form of missing data we do not consider in detail: data missing not-at-random (MNAR), where non-response is a result of \textit{unobserved} features of the DGP. One particular manifestation of this phenomenon may be where the CTR model has features that the post-click model does not have. This case is particularly problematic as it would not be possible to model these relationships in the post-click model. We leave this case, as any correction applied to AL sequences will be dependent on the missing relationship being congenial with the data observed. 

\subsection{Performance difference effects under MAR and MCAR}

In both the MAR \textit{and} MCAR contexts, fewer responses from the querying function will impair the performance of the model (holding constant the number of training rounds). In the MAR case, however, the presence of local regions of non-response can further impact the model by unbalancing the training data (relative to the full-response model). The presence of this additional source of degradation, therefore, suggests that AL performance can be \textit{worse} under MAR non-response compared to MCAR non-response. 

The extent of this divergence, we hypothesise, depends on how skewed the non-response distribution is, and thus the extent to which the model is able to explore certain areas of the feature space. In the extreme case, suppose that there are inaccessible regions of the data that, if queried, never return a label -- they act like ``black holes" that absorb the entire exploration budget without returning any labels. In this case, the imbalance effect will be large, because despite the high informativeness of this region, the model is totally prohibited from improving in this area. As a result, holding constant the unconditional rate of non-response, we would expect a large differential in model performance between MAR and MCAR missingness.

This case can be contrasted against one where there is still imbalance in the non-response distribution, but where there is nevertheless \textit{some} possibility of returning a label, and so some chance for model improvement. As the imbalance in the distribution of non-response lessens, and thus approaches the distribution under MCAR, then this differential should disappear. We demonstrate this expectation in the next section.

\section{Adjusting for the probability of non-response}
\label{sec:fix}

\textbf{General approach} In AL contexts without non-response, the ``utility" of a label is its informativeness to subsequent model training, i.e., $U_{y, t+1} = \mathcal{I}(\mathbf{x} \mid \mathcal{M}_t)$. For example, in uncertainty sampling contexts, the most informative labels are defined as those from regions of the feature space which the model is most uncertain about. When there is the possibility of (random) non-response, however, the query utility should be conditional on both the probability of response and the resulting label's informativeness.

Hence, we propose a simple adjustment to how query targets are selected, by optimizing the \textit{expected} utility (EU) of the informativeness score:
\begin{equation}
\label{eq:exp_util}
    EU = \sum_{\mathbf{x} \in \mathbf{X'}} \mathcal{I}(\mathbf{x} \mid \mathcal{M}_t) \times \text{P}(R_\Omega \mid \mathbf{x}), \ s.t. |\mathbf{X'}| = b,
\end{equation}
where $b$ is the total budgeted number of queries and $\mathbf{X'}$ are the query targets. Algorithm \ref{alg:exp_unc} details the implementation of this expected utility sampling strategy.\footnote{Following Lattimore \citep[p.418]{lattimore2020bandit}, we use the $\text{TopM}(\mathbf{a},m)$ operator to denote the largest $m$ values in vector $\mathbf{a}$. This step has complexity $O(n \log m)$ for choosing $m$ largest out of an array of length $n$, when implemented with a MinHeap. Note also line 5 sums the logged scores, which is equivalent to multiplying the estimates but more numerically stable.} Note that non-response is realised ``by nature".

\renewcommand{\algorithmiccomment}[1]{\bgroup\hfill//~#1\egroup}
\begin{algorithm}[t]
\caption{Expected Utility Active Learning}
\small

\hspace*{\algorithmicindent} \textbf{Input} $\mathbf{X}^\text{pool} \subset \mathbb{R}^{n \times k}$; $\mathbf{X}^\text{train} \subset \mathbb{R}^{m \times k}$; $\mathbf{y}^\text{train} \in \{0,1\}^{m}$; \\
\hspace*{\algorithmicindent} Target estimator $\mathcal{M}$ and non-response estimator $\mathcal{P}$; $n > 0$. \\
\hspace*{\algorithmicindent} \textbf{Output} $\mathcal{M}_T,$ a target model after $T$ rounds \\
\begin{algorithmic}[1]
\vspace{-1em}
\STATE $\mathcal{M}_0 \gets \mathcal{M}(\mathbf{X}^\text{train}, \mathbf{y}_\text{train})$

\FOR {t in 0:T}
    \STATE $\mathbf{u} \gets \mathcal{I}(\mathbf{X}^\text{pool}, \mathcal{M}_t)$  \COMMENT{Calculate informativeness scores}
    
    \STATE $\mathbf{p} = \mathcal{P}(\mathbf{X}^\text{pool})$ \COMMENT{Estimate response probabilities}
    
    \STATE $\mathbf{i} \gets \text{TopM}\big(\log(\mathbf{u}) + \log(\mathbf{p}), n\big)$ \COMMENT{Choose indices with highest expected utility}
    
    \STATE $\mathbf{X}^\text{query} \gets \mathbf{X}^\text{pool}[\mathbf{i}]$ 
    
    \STATE $\mathbf{X}^\text{train} \gets \mathbf{X}^\text{train} \cup \mathbf{X}^\text{query}[\mathbf{r}_t^\text{query}=1]$ \COMMENT{Add successful queries to training data} 
    
    \STATE $\mathbf{y}^\text{train} \gets \mathbf{y}^\text{train} \cup \mathbf{y}^\text{query}[\mathbf{r}_t^\text{query}=1]$

    \STATE $\mathcal{M}_{t+1} = \mathcal{M}_t(\mathbf{X}^\text{train}, \mathbf{y}^\text{train})$ \COMMENT{Update model state}
\ENDFOR
\end{algorithmic}
\label{alg:exp_unc}
\end{algorithm}

By incorporating the cost of querying non-responsive regions of the feature space, this adaptation should prevent the model from ``wasting" its budget on areas where labels are informative but the likelihood of observing one is very small.

\begin{figure*}[t]
    \centering
    \includegraphics[width = \textwidth]{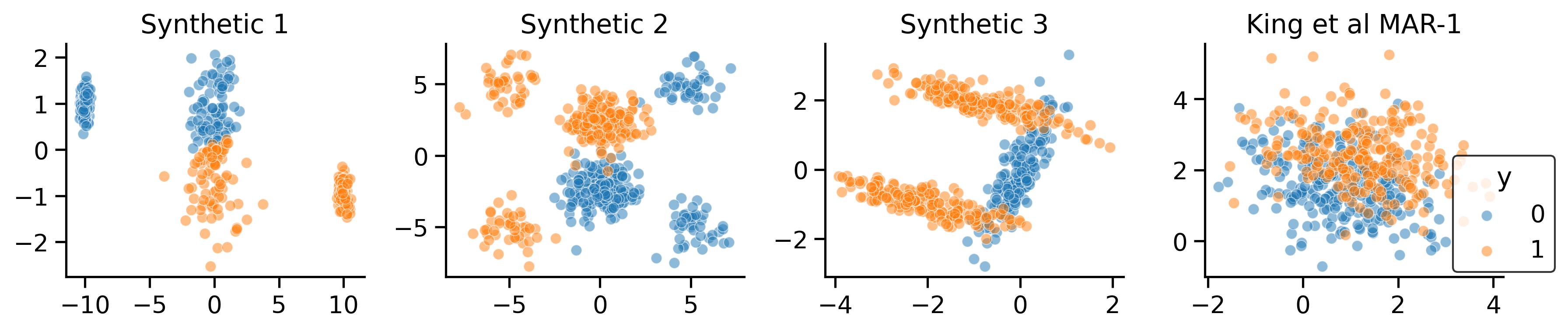}
    \vspace{-0.4cm}
    \caption{Illustration of synthetic datasets used in AL experiments. Note: MAR-1 is a restricted view of the data, and only shows two of the five X dimensions.}
    \label{fig:dgp}
\end{figure*}

The economic aspect of this correction has one further advantage: with sufficient training steps, the model will saturate in regions of the feature space where responses are likely. In these instances, the informativeness (however defined) will be so minimal that, despite high non-response rates, the model is able to switch to sampling solely from highly missing regions.

By design, therefore, this adjustment still grants the AL model leeway to explore regions of the feature space: as the model becomes more confident over regions with higher response rates, it becomes useful, and relatively less costly, to sample from low response regions. Model progress at this point will, clearly, be slow, but it may yield informative examples with a sufficient budget (and willingness to pay).

\textbf{Non-response prediction error} In practice, calculating the expected utility involves an \textit{estimate} of the probability of response. In many settings, it may be possible to train or develop such a model before the active learning process. Note, for the purpose of training a CTR model, impressions without a click have a (negative) label. Therefore, the CTR model has more data available than the post-click model that we target in active learning. Hence, the CTR model may be accurate in regions where the post-click model is not.

That said, the non-response probabilities from such a model are nevertheless estimates. In some cases, these estimates may have (considerable) uncertainty. We may therefore want to explore these areas optimistically. This trade-off will be especially important, for example, where the non-response model is trained on a small number of observations, or where the non-response model is retrained iteratively during AL. We can address this issue by replacing the predicted probability of response $\mathbf{p}$ in Algorithm \ref{alg:exp_unc} with:
$$
\mathbf{p} = \mathcal{Q}(\mathbf{X}^{pool}, 0.95),
$$
where $\mathcal{Q}(\cdot, 0.95)$ returns the 95th quantile of the distribution of predictions for each $\mathbf{x} \in \mathbf{X}^\text{pool}$. This strategy resembles upper confidence bound (UCB) sampling, a common bandit algorithm. It requires an estimator that is capable of returning an (approximate) posterior distribution, like a Gaussian Process model or bootstrapped ensemble. For well-trained (i.e., precise) non-response models, note also that we would expect (and find) limited changes incorporating UCB, as the 95th quantile will be close to the predicted probability.

We refer to the final correction as the ``\textit{Upper Confidence Bound of the Expected Utility}" (UCB-EU). One major advantage of this strategy is that it is deliberately compatible with different base AL sampling strategies. Plausibly, any informativeness metric $\mathcal{I}$ and non-response estimator $\mathcal{P}$ can be plugged into Algorithm \ref{alg:exp_unc}. As a result, the overall complexity class depends on the choices of $\mathcal{I}$ and $\mathcal{P}$. In our experiments, we demonstrate this approach using QbC, uncertainty, and random sampling strategies. 

\section{Empirical evidence of AL model degradation under non-response}
\label{sec:sims_nf}

\begin{figure*}[t]
    \centering
    \vspace{-0.1cm}
    \includegraphics[width = \textwidth]{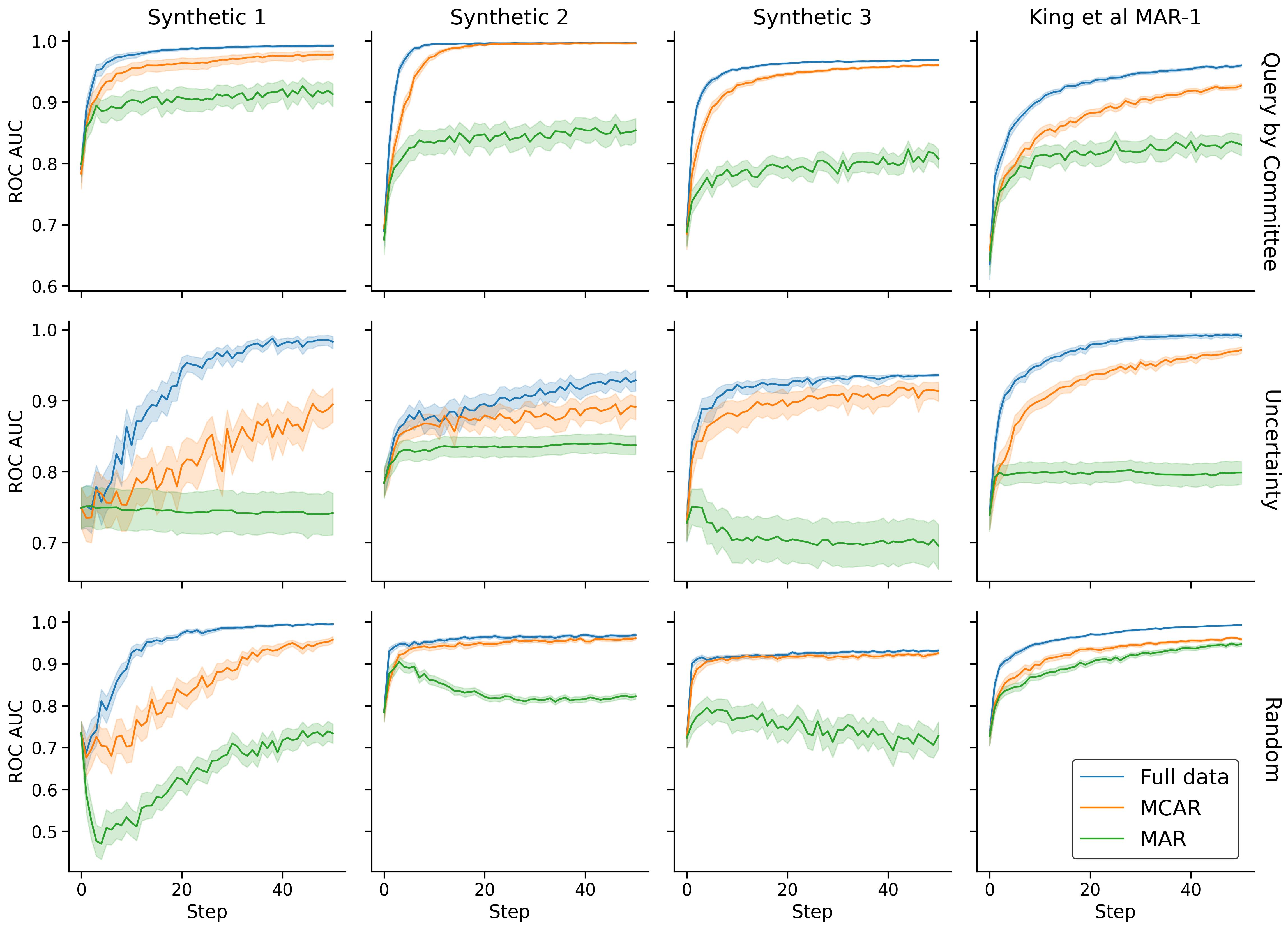}
    \vspace{-0.5cm}
    \caption{AL model performance in the presence of non-response, using different sampling strategies. $\mathbb{E}[R] = 0.3$ across all simulations. Observations in the missing region had a 0.001 probability of response. Shaded areas show the 95\% confidence interval over 200 separate simulations (per non-response mechanism).}
    \label{fig:curtailed_sims}
    \vspace{-0.2cm}
\end{figure*}

\subsection{Experimental setup}
\textbf{Synthetic data} We focus on four synthetic data scenarios, as illustrated in Figure \ref{fig:dgp}. Synthetic 1 simulates a linear decision boundary with clusters offset in one dimension, such that $\mathbb{E}[Y] = 0.1$. Synthetic 2 and 3 are based on DGPs presented in \citep{huang2014representative}, which have been demonstrated to prove challenging for AL strategies like uncertainty sampling \citep{yang2018benchmark}. In Synthetic 2, the data contain six normally distributed clusters in two dimensions, with $\mathbb{E}[Y] = 0.5$. In Synthetic 3, the data are distributed in a rotated U-shape where the distribution of positive cases intersects the tails of the two other sides. Finally, King et al's MAR-1 dataset \citeyearpar{king2001analyzing} considers a multivariate normal distribution with moderate correlations between five dimensions, based on a simulation design used in inference-focused missing data studies \citep{king2001analyzing,lall2022midas}. We convert this final scenario into a classification problem by implementing a non-linear decision boundary: $y_i \xleftarrow{} 5X_0 - 4 X_1 + 3X_2 - 2X_3 + X_4 + 0.5X^2_0 + 3X_1X_2 >= c$, setting $c$ such that $\mathbb{E}[Y] = 0.1$.

\textbf{Non-response mechanisms} As noted in Section \ref{sec:theory}, we consider three non-response mechanisms. First, that there is no non-response and so we observe the full data (i.e., $R_\Omega=1$ across the entire feature space). This mechanism is the benchmark, or ideal, data generating context and yields the full-response model. Second, that the data is MCAR. This mechanism helps understand the volume effect of training (relative to the full data mechanism), since there is no correlation between the probability of non-response and the feature space. To model MCAR missingness, we randomly induce non-response uniformly across the feature space. Third, that the data is MAR, where missingness is a function of the observed data. In our experiments, we partition each DGP's feature space into two regions at some threshold value along its first explanatory dimension. Therefore, the missingness is correlated with the value of this dimension. On one side of this threshold, the ``low response region", we impose a high probability of \textit{non}-response, $P(R_\Omega =1) = 0.001$, and on the other side we impose a high probability of response. We hold constant the unconditional missingness probability (to match the MCAR mechanism), by adjusting the specific threshold for the two missing regions.\footnote{Given some defined probability of response in the first region, $p_1$, and targeted unconditional response probability $p*$, we assign a response probability $p_2 = 1 - p_1(1 - p*)$ to the other region.}

\textbf{Simulations} For each scenario and non-response mechanism, we simulate 50 rounds of AL. Each model is seeded with two random examples. In each round, the model queries the 10 most informative unlabelled examples. We repeat each simulation 200 times to calculate the expected performance and 95\% confidence intervals for each step of training, assessed using the ROC-AUC score on 1000 holdout examples. 

We run identical versions of the experiment using QbC, uncertainty, and random sampling strategies. QbC is a commonly used query strategy, and it largely addresses documented deficiencies of simpler methods like uncertainty sampling \citep{settles2009active}. Our QbC model uses a random forest classifier as the ensemble. We also benchmark uncertainty sampling as a simple form of AL strategy. The uncertainty sampler uses a linear support vector machine (SVM) as the learning algorithm. Finally, we include random sampling as a naive acquisition strategy to demonstrate the generality of our correction.\footnote{All simulations were run on a single server instance, using GPU processing equivalent to an NVIDIA Tesla P100 GPU with 16GB memory. The total time taken to run 200 simulations for every combination of missing mechanism, strategy, and DGP was 12 hours. The non-response model for each DGP took approximately 5 minutes to train.} 

\textbf{Non-response corrections} We implement our proposed algorithm, for each AL strategy, as follows:
\begin{itemize}
    \item For \textit{QbC} sampling, we use McCallum et al.~\citep{mccallum1998employing}'s modified acquisition function, which is based on the log of the maximum Kullback-Leibler divergence for each label to the log of the UCB predicted probability of response
    \item For \textit{uncertainty} sampling, we add the log entropy of the predictions over the pool examples to the log of the UCB predicted probability of response
    \item For \textit{random} sampling, we generate the UCB predicted upper confidence bound and softmax these values so the scores sum to 1. $n$ new observations are then randomly selected according to this vector of sampling probabilities
\end{itemize}

To model the probability of non-response, and to perform UCB sampling of a posterior, we pre-train Gaussian Process (GP) models for each simulation DGP.\footnote{We implemented the GP models using \texttt{GPyTorch}~\citep{gardner2018gpytorch}. We fit a constant prior mean function and scaled RBF kernel for the covariance matrix, using a Bernoulli Likelihood function. Since our MAR mechanism is relatively simple, we find these models prove to be very effective classifiers: all models have ROC-AUC values above 0.995 and mean absolute errors at or below 0.04.}

\subsection{Results}
\label{sec:mar_curtailed}

We first consider the uncorrected impact of non-response on AL performance (Section \ref{sec:uncorrected}). We then demonstrate how our solution improves on these results (Section \ref{sec:corrected}), before exploring in more detail cases where non-response poses particularly acute problems for active learning (Section \ref{sec:why}).

\subsubsection{Uncorrected impacts of non-response}\label{sec:uncorrected}

Figure \ref{fig:curtailed_sims} plots the simulation results without any correction. Across all four synthetic DGPs, we see that it is the MAR form of missingness that is more harmful to active learning than other non-response mechanisms. Over sequential training steps, the MAR-affected model either fails to improve or its performance even \textit{deteriorates} as we collect more data. This effect is particularly clear in the Synthetic 3 scenario, where the inability to query parts of the feature space biases the model and leads to worse performance over sequential training, under both uncertainty and random sampling.

Across all scenarios, sequential steps yield model improvements under MCAR non-response, but the extent of the difference between this model and the full data model, on average, differs considerably. We do observe the effect of having a lower volume of training data under both the Synthetic 1 and MAR-1 DGPs, with MCAR performance significantly worse than the full data model. This difference is less pronounced under QbC, which is relatively less affected by MCAR non-response. In Synthetic 3, MCAR performance is substantially poorer, by 50 steps all three sampling strategies see similar performance to the full data model, although there are distinguishable differences earlier in training.

\begin{figure*}[t]
    \centering
    \includegraphics[width = 0.95\linewidth]{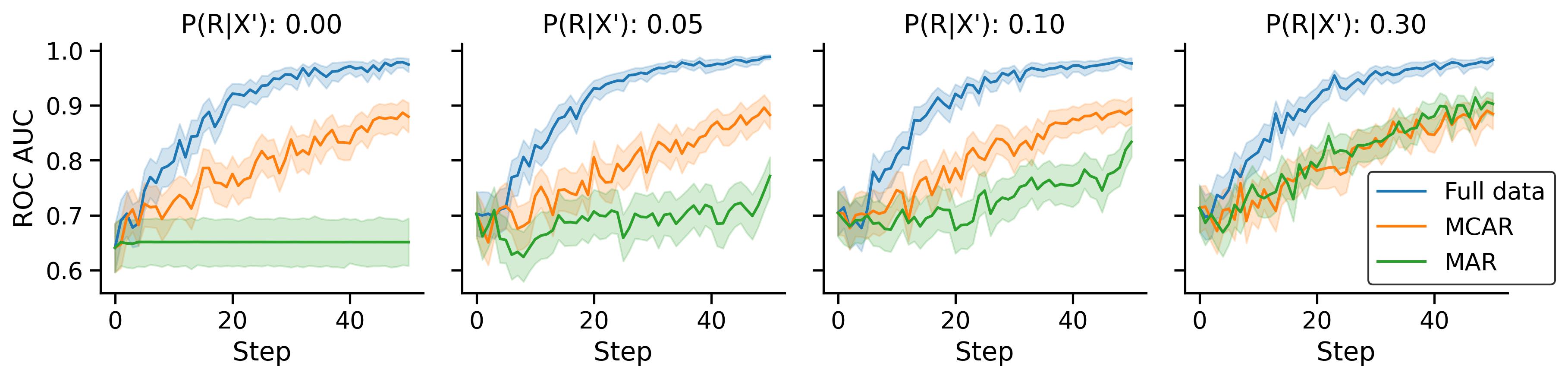}
    \vspace{-0.15cm}
    \caption{The effect of imbalance on model performance between MAR and MCAR non-response mechanisms. The probabilities above each panel indicate the probability of response in the low response region of the feature space. Shaded areas show 95\% confidence intervals over 200 simulations.}
    \label{fig:mar_gap}
\end{figure*}

We also hypothesised that the difference in model performance between MAR and MCAR non-response mechanisms results from an imbalance effect. In the most extreme case, when regions of the feature space preclude label acquisition entirely, AL strategies simply cannot learn about these regions and thus may perform poorly.

We test this hypothesis by adjusting the probabilities of non-response either side of the non-response threshold. To hold constant the volume effect inherent to increased missingness, we adjust the cut-off position to maintain the same unconditional probability of response ($\mathbb{E}[R] = 0.3$). We run 200 simulations per missingness mechanism for each of the five different non-response probability tuples. Where the severity of the non-response threshold is most acute (i.e., fully observed on one side, and never-observed on the other) we would expect the largest difference in performance. Conversely, as the severity of this threshold decreases, such that the non-response probabilities approach uniformity, we would expect MCAR and MAR models to converge in performance.

Figure \ref{fig:mar_gap} reports the average results over these simulations, using an uncertainty sampling strategy. The MCAR-afflicted model improves (slowly) because the missingness probability is constant across the \textit{entire} feature space. By contrast, under MAR missingness it is \textit{impossible} to learn about some portion of the feature space. This becomes increasingly pronounced for low values of $P(R_{\Omega}=1 \mid \mathbf{X})$ in the low-response region, in which case it results in flat learning curves and large gaps between MAR and MCAR. As we increase $P(R_{\Omega}=1 \mid \mathbf{X})$ the MAR/MCAR gap decreases, such that in the final panel, where the probability of response in the low response region is the same as the marginal rate of response, MAR and MCAR performance are similar.

\subsubsection{MAR impacts with UCB-EU correction}
\label{sec:corrected}

\begin{figure*}[t]
    \centering
    \includegraphics[width = 0.95\linewidth]{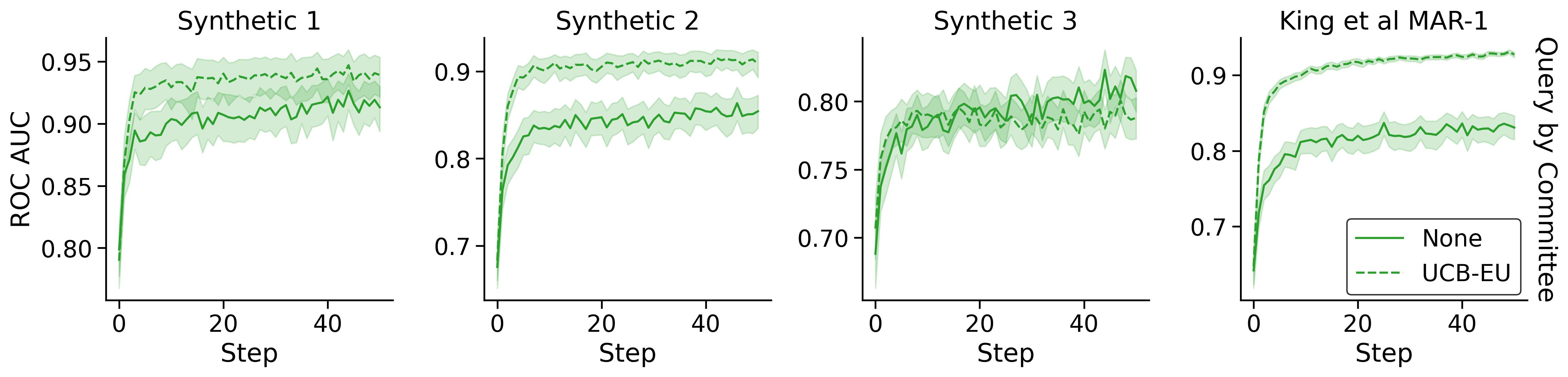}
    \vspace{-0.15cm}
    \caption{Comparison of Query-by-Committee AL performance with and without UCB-EU correction, under MAR non-response. The DGP is identical to the results presented in Figure \ref{fig:curtailed_sims}.}
    \label{fig:qbc_fix}
\end{figure*}

Compared to the uncorrected results, the modified algorithm shows better performance for many combinations of strategy and DGP. As shown in Figure \ref{fig:qbc_fix}, implementing a cost to searching low response regions improves the ability of the QbC strategy to refine its selection of unlabelled examples to query. In all but the Synthetic 3 DGP, using the UCB-EU correction leads to substantially better model performance. 

\begin{figure*}[t]
    \centering
    \includegraphics[width = 0.95\linewidth]{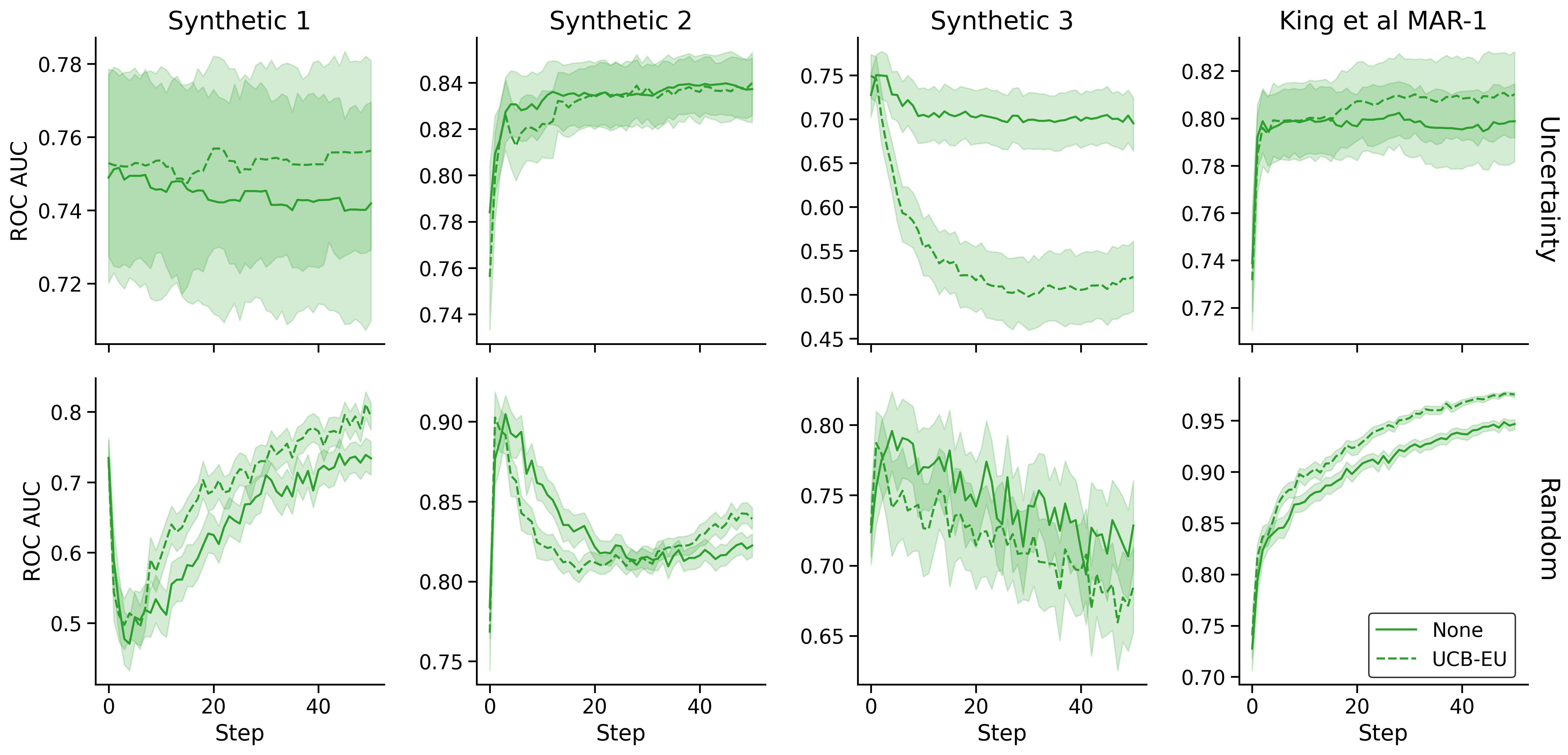}
    \vspace{-0.15cm}
    \caption{Comparison of Uncertainty and Random AL performance with and without UCB-EU correction, under MAR non-response. The DGP is identical to the results presented in Figure \ref{fig:curtailed_sims}.}
    \label{fig:other_fix}
\end{figure*}

Figure \ref{fig:other_fix} plots the results of using the UCB-EU correction for uncertainty and random sampling strategies, respectively. In the case of random sampling, the UCB-EU correction also yields considerably better model performance, although the naivety of the baseline strategy appears to add additional noise to the learning process early in training. By around 30 steps of AL training, and similar to the more performant QbC strategy, the UCB-EU corrected models outperform the baseline model under the Synthetic 1, 2, and MAR-1 DGPs. 

In the case of uncertainty sampling, the performance improvements are less pronounced. However, we observe that uncertainty sampling consistently underperforms random sampling on these datasets. This result is not surprising given existing literature: it is widely documented that uncertainty sampling does not always outperform random sampling \citep{attenberg2011inactive,settles2012uncertainty,yang2018benchmark,jin2022cold,TifreaCY2}. We cannot expect UCB-EU non-response corrections to make a sampling strategy sample efficient under non-response settings, if that sampling strategy is not sample efficient without non-response to start with. Therefore, we believe this is failure of uncertainty sampling itself rather than a failure of the UCB-EU adjustment.
\renewcommand{\floatpagefraction}{.9}%
\subsection{Where should the model query?}
\label{sec:why}
MAR non-response on Synthetic 3 leads to \textit{worse} performance under our modified algorithm. This result is most pronounced in the case of uncertainty sampling. We conjecture that the degradation in performance observed under the Synthetic 3 DGP is a result of non-response forcing the model to fit on a low non-response part of feature subspace that is non-representative of the general population (and hence, of the test set distribution). In other words, modifying the AL sequence to discount the utility of low-response regions may narrow the model's focus too much, therefore learning a ``local" decision boundary that is optimised only for the low non-response parts. This boundary is quite different from the global optimal boundary that would result in performance similar to the full data model.

We can illustrate this point in two ways. First, we abstract away from AL and simply assume areas of the feature space are fully observed ($R_{\Omega(\mathbf{x})} = 1$) or never observed ($R_{\Omega(\mathbf{x'})} = 0$). We  take a large $N$ sample and train a target model on this data, which should approximate the long-run performance of the AL model with many queries from the pool. We conduct this exercise four times, varying where the non-response threshold is applied, which has the effect of varying the proportion of the DGP that is observable.
\looseness=-1
\begin{figure*}[t]
    \centering
    \vspace{-0.1cm}
    \includegraphics[width=0.95\textwidth]{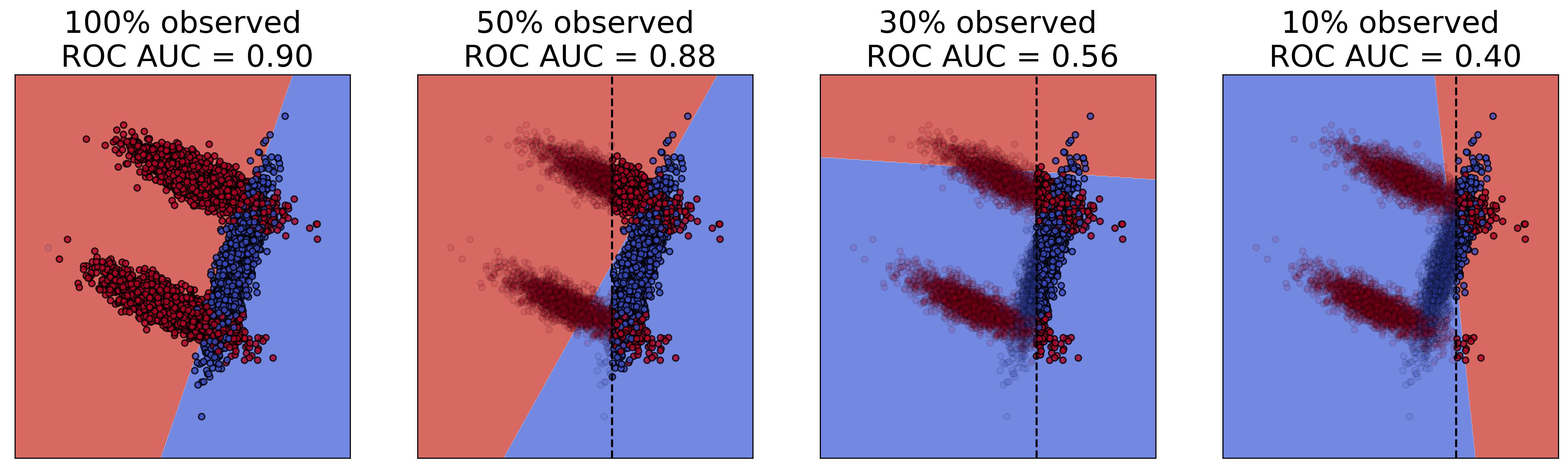}
    \vspace{-0.1cm}
    \caption{The effect of non-response on estimated decision boundaries in Synthetic 3 DGP. Dashed vertical lines indicate the threshold between low-response (left) and full-response (right) regions of the covariate space. Blurred datapoints indicate specific examples of the underlying DGP that are not available to the model during fitting. All models are trained on 6000 observations, to minimise uncertainty over the decision boundary.}
    \label{fig:change_boundary}
    \vspace{-0.2cm}
\end{figure*}

Figure \ref{fig:change_boundary} plots the DGP and the estimated decision boundaries in each case. The leftmost panel displays the optimal linear decision boundary with zero non-response. The dashed vertical line indicates the boundary between the low-response (left) and full-response (right) regions of covariate space. While a linear boundary cannot perfectly distinguish labels under this DGP's distribution, the ROC AUC score is nevertheless high with a 100\% response rate. As the response rate decreases, the decision boundary rotates as it fits to the narrower set of points in the high-response region on the right side of the dashed line. The optimal decision boundary, conditional on the observed data, becomes increasingly different from the ``true" decision boundary as the proportion of non-response increases. By the third panel, the decision boundary is mis-classifying the entire bottom wing of the DGP distribution. By the final panel, moreover, the classification boundary has inverted.  Unsurprisingly, the ROC AUC scores decline substantially as non-response affects more of the area. These results confirm our conjecture that in the ``ideal" world where an AL sequence focuses only on \emph{observable} examples, this myopia can lead UCB-EU to exacerbate the selection bias effects that are always present when using active learning. When response probabilities become infinitesimally small, the harm from this selection bias may outweigh the benefits of UCB-EU that are obtained from increased annotation volume relative to uncorrected AL.

\begin{figure}{t}{6cm}%[t]
    \centering
    \vspace{-0.4cm}
    \includegraphics[width = 0.95\textwidth]{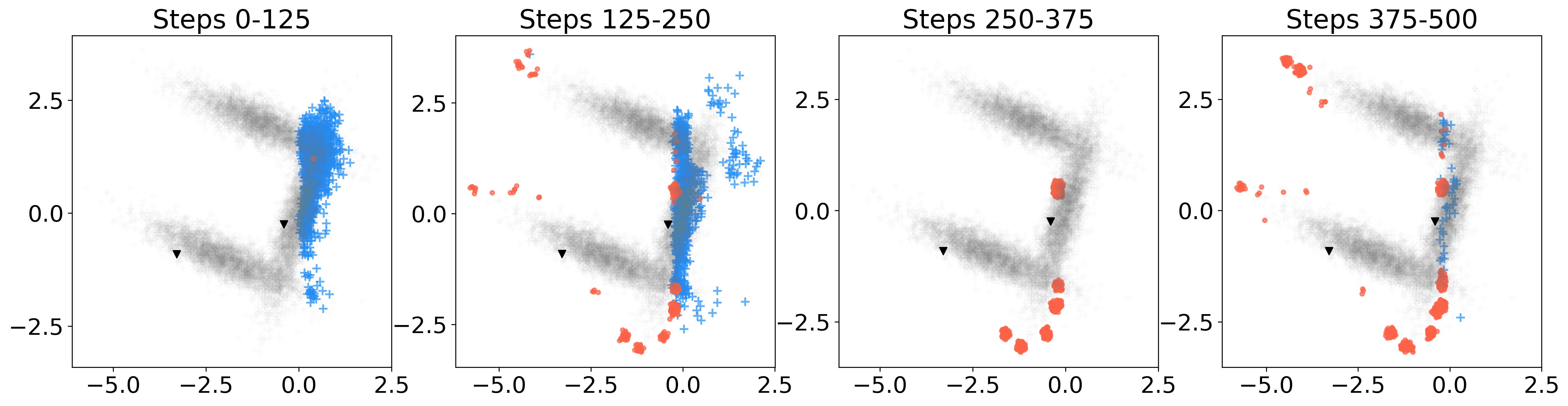}
    \vspace{-0.2cm}
    \caption{Query history using the QbC strategy and UCB-EU correction with Synthetic 3 DGP. The model is trained for 500 steps with a batch size of 10 and two initial labelled examples (black daggers). The underlying DGP is shown as a faint distribution of grey dots in the background. Blue crosses indicate the queried label was observed, and red circles indicate non-response. Each facet shows the (binned) queries within the sequence of AL iterations.}
    \label{fig:query_time}
    \vspace{-0.1cm}
\end{figure}

Next, we verify this expectation by tracking the query targets in our AL experimentation setup, using the UCB-EU correction and QbC sampling strategy. Figure \ref{fig:query_time} plots each individual query attempt and whether the resulting label was observed. We bin the AL iterations into sequential facets, to show \textit{when} in the sequence each query was made. In early iterations of training, the AL model pays almost exclusive attention to the high response region of the feature space. In particular, in steps 1 to 125, it focuses on the top-right section (right of where the dashed line was). This is the result of the UCB-EU correction: when the model has few observations, the most informative queries, and those with the highest response probabilities, are in this region. As the model reaches the point where it has exhausted the learning potential in this region, it switches its focus to attempt to learn in the low-response parts of covariate space where the model does not yet have much training data. Consistent with Figure \ref{fig:change_boundary}, we now observe selection bias induced by non-response. The UCB-EU correction prevents the AL sequence from ``wasting" early queries on parts of the feature space where the probability of getting a valid label are very small, but which would ultimately yield a different (and better) decision boundary if observed. 
\section{Case study: Taobao shopping behaviour modelling}
\label{sec:taobao}
Consider an e-commerce platform that ranks products that are presented as a list of items on its homepage. A click on an item in this ranking brings the user to a product details page with a checkout button. The aim is to rank products in a way that maximises the conversions on the platform. Imagine that the product ranking system is designed to factorize the estimation task of conversion probabilities into two separate machine learning models: $\mathit{P}(\mathit{conversion}) = \mathit{P}(\mathit{click}=1) \mathit{P}(\mathit{conversion} \mid \mathit{click}=1)$, i.e., a \emph{CTR} model that estimates the probability that a user will click on a product in the ranking, and a so-called \emph{post-click} model that estimates the probability that a user who clicked on the product and lands on the product details page will proceed to purchase the item. Factorization into separate CTR and post-click models is common in the e-commerce and online advertising industry~\citep{ma2018entire,lin2023tree, barbieri2016improving,rosales2012post}. The resulting quantity $\mathit{P(conversion})$ yields the probability that a user will purchase an item in the ranking, and serves as our sorting criterion for ranking purposes. The CTR model can be trained on a dataset of all items that users viewed, while the post-click model can only be trained on a data set of page loads of the product details page (hence, on clicked items).

Imagine a setting where we aim to improve the \emph{post-click} model component through AL exploration, and to achieve this exploration we can intervene in what products we display in the product ranking. In this setting, whether a user clicks the item or not constitutes the non-response mechanism, i.e., we do not obtain a label for the post-click model if the user does not click.\footnote{The absence of a click should be interpreted as missing data for the $P(\mathit{conversion}\mid\mathit{click}=1)$ estimator and not as a negative label. Including non-clicked impressions as negative labels would result in a direct estimation of the marginal $P(\mathit{conversion})$. Our aim here, instead, is to estimate the conditional probability $P(\mathit{conversion}\mid\mathit{click}=1)$ and then estimate $P(\mathit{conversion})$ by multiplying the conditional quantity by the estimated CTR, as is common in industry.}

We use the large-scale Taobao \citep{taobao} dataset that consists of product impressions, clicks, and conversions like in the setup that we described above. We set out to assess whether our proposed method helps correct for the potential non-response bias in this context. This data includes a random sample of 1.1 million users and their corresponding behaviours on the platform between 6-13 July 2017 ($n{=}700$ million behaviour logs). We use a combination of the user, product, and behaviour data so that we can record which users clicked which product (the labelling process) and which users purchased the product (the post-click target model).

To simulate the AL process, we initialise a random forest model with 50 random observations. We conduct 25 steps of AL, and at each step query 5000 new examples from the pool ($n{=}12.3$ million user-product-behaviour triples), to simulate impressions by users. We realise non-response using the \textit{observed} CTR indicator from the data. At the end of each step, \textit{all} queried observations are removed from the pool to mimic the temporary nature of the impression. We also test a version of this simulation, approximating the re-querying strategy of \citet{yan_2015}, where queried observations with non-response are replaced in the pool rather than removed. We holdout a test set of 10,000 observations (\textit{post-click} user-item pairs). 

Finally, to assess how the performance of the non-response (i.e., CTR) model affects the improvement produced by UCB-EU, we run our simulation multiple times, implementing a series of ``synthetic'' CTR models where we deliberately vary the model's ROC-AUC score across simulations. These CTR models use, as their base prediction, the true click-through scores (from the pool). We then corrupt this vector, by inverting $1-\text{ROC-AUC}$ labels randomly, to generate CTR models of varying performance. We compare our general method to the average case Bayesian Active Learning with Abstention Feedbacks (BALAF) proposed by \citet{NGUYEN2022242}. This approach uses $L_2$-regularised logistic regression models to approximate learning the posterior of the target and non-response models, using \textit{maximum a posteriori} estimation, and we pre-train the missing model using the entire pool.

\begin{figure}[t]
\vspace{-0.25cm}
    \centering
    \includegraphics[width = 0.95\textwidth]{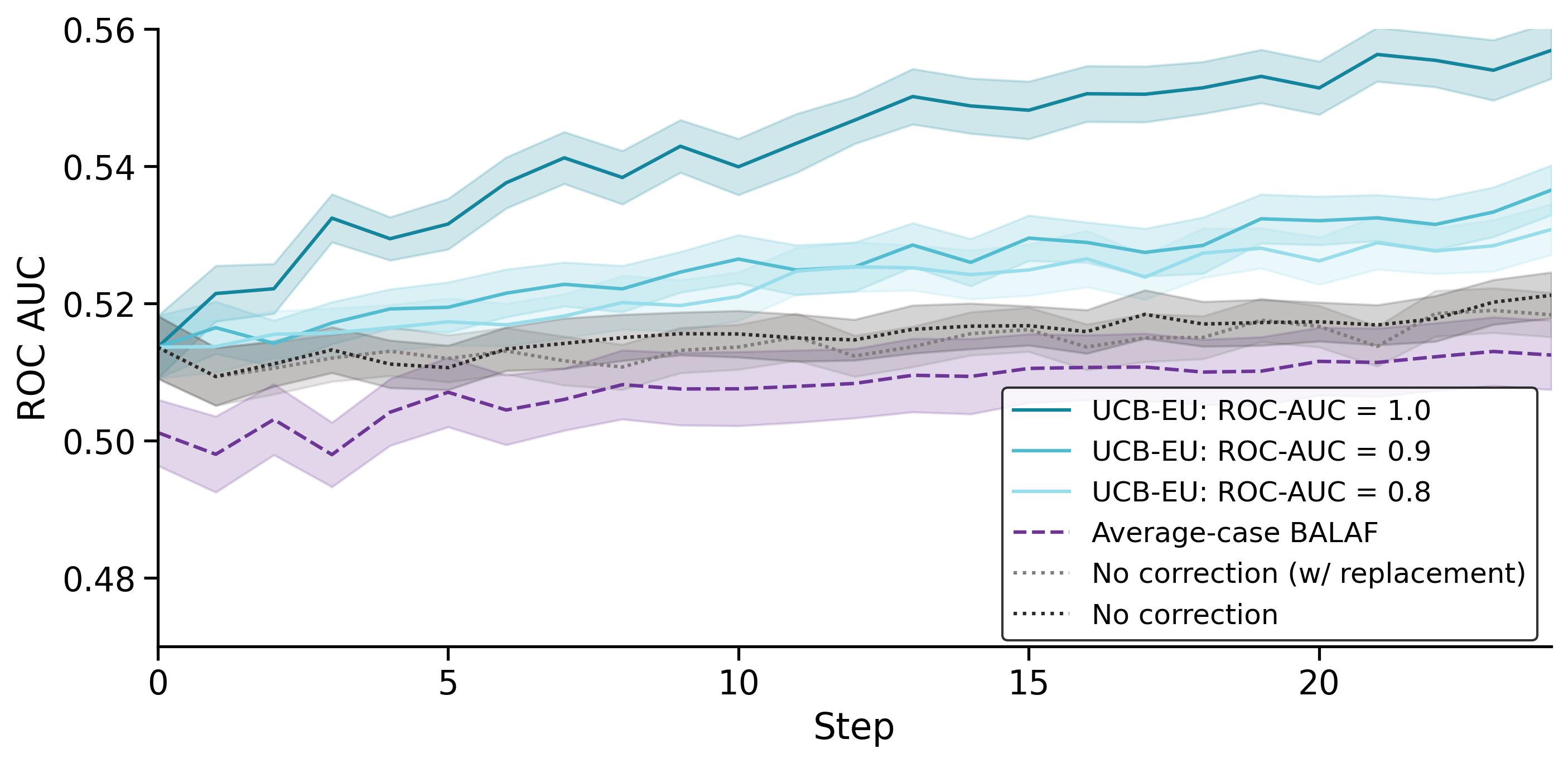}
    \vspace{-0.2cm}
    \caption{Target model performance of UCB-EU AL-trained models compared to other algorithms. Lines plot the mean ROC-AUC at each step and the shaded areas plot the 95\% confidence intervals.}
    \label{fig:taobao}
\vspace{-0.4cm}
\end{figure}

Figure \ref{fig:taobao} displays the results of each scenario, averaged over 100 independent simulations of each AL process. Compared to the uncorrected QbC model, using UCB-EU for 25 steps of AL can yield up to a 10\% improvement in the model's ROC-AUC relative to AL without the UCB-EU correction. However, the quality of the non-response model matters: less accurate non-response models result in smaller gains from UCB-EU as the adjustments become noisy.

Replacing queried datapoints has very little impact on model performance, and is still markedly worse than the UCB-EU correction. The average-case BALAF strategy results are also largely indistinguishable from the uncorrected AL model as the confidence intervals largely overlap. We believe that this result is unsurprising: as the authors themselves note, the performance of BALAF breaks down when the rate of non-response is high.
\vspace{-0.2cm}
\section{Discussion \& Conclusion}
Non-response in the feature space limits model performance. In this paper, we extend our understanding of this process by demonstrating that the \textit{form} of non-response, and its potentially \textit{correlated} relation with the underlying DGP, leads to differential impacts on model performance, and can even degrade model performance over time. Biased non-response of this type may be particularly common where AL involves user interactions. We demonstrate that we can mitigate this loss in performance by adjusting the utility of querying labels by the estimated probability of non-response.

Importantly, we find, however, that non-response distributions can lead to local decision boundaries inconsistent with the theoretically global optima. This is a challenging problem for AL methods, which deserves attention in future research.
It is also worth noting that, at any given step, our correction may yield labels that are less intrinsically informative but which avoid the model learning nothing by receiving a null label. Therefore, while our approach should have improved performance relative to a naive AL model, it faces the hard constraint that non-response may ultimately curtail the informativeness of the data available to the model.

Finally, our experiments use a high non-response probability. This severity creates a sharp threshold in the feature space, and likely contributes to the limited effect our correction has in some contexts (relative to random non-response). As we demonstrate, separately, where the probability of response is higher (even under MAR conditions), the imbalance effect on model performance becomes relatively less substantial. We do, however, think that contexts of high (or even total) non-response are common in areas where, for example, labelling requires costly user behaviour.

\vspace{-0.2cm}
\section*{Declarations}
\noindent \textbf{Ethical Statement} There are no ethical issues. This research was considered and approved by the London School of Economics' research ethics procedure (ref. 183654).

\noindent \textbf{Funding} No funding was received to assist with the preparation of this manuscript.

\noindent \textbf{Competing interests} The authors have no relevant financial/non-financial interests.

\noindent \textbf{Data Availability} All simulation data-generating processes are described in the paper. All data used in Section \ref{sec:taobao} is available at \url{https://tianchi.aliyun.com/dataset/56}.\looseness=-1

\setlength{\bibsep}{8pt plus 0.2ex}
\vspace{-0.2cm}
\bibliography{refs_full}

\begin{thebibliography}{41}
\providecommand{\natexlab}[1]{#1}
\providecommand{\url}[1]{{#1}}
\providecommand{\urlprefix}{URL }
\providecommand{\doi}[1]{\url{https://doi.org/#1}}
\providecommand{\eprint}[2][]{\url{#2}}
 \bibcommenthead

\bibitem[{Amin et~al(2021)Amin, DeSalvo, and Rostamizadeh}]{amin2021}
Amin K, DeSalvo G, Rostamizadeh A (2021) Learning with labeling induced
  abstentions. In: Advances in Neural Information Processing Systems, pp
  12576--12586

\bibitem[{Attenberg and Provost(2011)}]{attenberg2011inactive}
Attenberg J, Provost F (2011) Inactive learning? difficulties employing active
  learning in practice. ACM SIGKDD Explorations Newsletter 12(2):36--41

\bibitem[{Audibert et~al(2010)Audibert, Bubeck, and Munos}]{audibert2010best}
Audibert JY, Bubeck S, Munos R (2010) Best arm identification in multi-armed
  bandits. In: COLT, pp 41--53

\bibitem[{Barbieri et~al(2016)Barbieri, Silvestri, and
  Lalmas}]{barbieri2016improving}
Barbieri N, Silvestri F, Lalmas M (2016) Improving post-click user engagement
  on native ads via survival analysis. In: Proceedings of the 25th
  International Conference on World Wide Web, pp 761--770

\bibitem[{Bart{\'o}k et~al(2014)Bart{\'o}k, Foster, P{\'a}l, Rakhlin, and
  Szepesv{\'a}ri}]{bartok2014partial}
Bart{\'o}k G, Foster DP, P{\'a}l D, et~al (2014) Partial
  monitoring—classification, regret bounds, and algorithms. Mathematics of
  Operations Research 39(4):967--997

\bibitem[{Carcillo et~al(2018)Carcillo, Le~Borgne, Caelen, and
  Bontempi}]{carcillo2018streaming}
Carcillo F, Le~Borgne YA, Caelen O, et~al (2018) Streaming active learning
  strategies for real-life credit card fraud detection: assessment and
  visualization. International Journal of Data Science and Analytics 5:285--300

\bibitem[{Cortes et~al(2018)Cortes, DeSalvo, Gentile, Mohri, and
  Yang}]{cortes2018online}
Cortes C, DeSalvo G, Gentile C, et~al (2018) Online learning with abstention.
  In: International conference on machine learning, pp 1059--1067

\bibitem[{Elahi et~al(2016)Elahi, Ricci, and Rubens}]{elahi2016}
Elahi M, Ricci F, Rubens N (2016) A survey of active learning in collaborative
  filtering recommender systems. Computer Science Review 20:29--50

\bibitem[{Fang et~al(2012)Fang, Zhu, and Zhang}]{fang2012active}
Fang M, Zhu X, Zhang C (2012) Active learning from oracle with knowledge blind
  spot. In: Twenty-Sixth AAAI Conference on Artificial Intelligence

\bibitem[{Farquhar et~al(2021)Farquhar, Gal, and
  Rainforth}]{farquhar2021statistical}
Farquhar S, Gal Y, Rainforth T (2021) On statistical bias in active learning:
  How and when to fix it. arXiv preprint arXiv:210111665

\bibitem[{Freund et~al(1997)Freund, Seung, Shamir, and
  Tishby}]{freund1997selective}
Freund Y, Seung HS, Shamir E, et~al (1997) Selective sampling using the query
  by committee algorithm. Machine learning 28(2-3):133

\bibitem[{Gardner et~al(2018)Gardner, Pleiss, Weinberger, Bindel, and
  Wilson}]{gardner2018gpytorch}
Gardner J, Pleiss G, Weinberger KQ, et~al (2018) G{P}y{T}orch: {B}lackbox
  matrix-matrix {G}aussian process inference with {GPU} acceleration. In:
  Advances in neural information processing systems

\bibitem[{Hansen and Hurwitz(1946)}]{hansen1946problem}
Hansen MH, Hurwitz WN (1946) The problem of non-response in sample surveys.
  Journal of the American Statistical Association 41(236):517--529

\bibitem[{Huang et~al(2014)Huang, Jin, and Zhou}]{huang2014representative}
Huang SJ, Jin R, Zhou ZH (2014) Active learning by querying informative and
  representative examples. IEEE Transactions on Pattern Analysis and Machine
  Intelligence 36(10):1936--1949

\bibitem[{Jin et~al(2022)Jin, Yuan, Li, Wang, Wang, and Song}]{jin2022cold}
Jin Q, Yuan M, Li S, et~al (2022) Cold-start active learning for image
  classification. Information Sciences 616:16--36

\bibitem[{King et~al(2001)King, Honaker, Joseph, and
  Scheve}]{king2001analyzing}
King G, Honaker J, Joseph A, et~al (2001) Analyzing incomplete political
  science data: An alternative algorithm for multiple imputation. American
  political science review 95(1):49--69

\bibitem[{Lall and Robinson(2022)}]{lall2022midas}
Lall R, Robinson T (2022) The midas touch: Accurate and scalable missing-data
  imputation with deep learning. Political Analysis 30(2):179--196

\bibitem[{Lattimore and Szepesv{\'a}ri(2020)}]{lattimore2020bandit}
Lattimore T, Szepesv{\'a}ri C (2020) Bandit algorithms. Cambridge University
  Press

\bibitem[{Lewis(1995)}]{lewis1995}
Lewis DD (1995) A sequential algorithm for training text classifiers:
  Corrigendum and additional data. In: ACM SIGIR Forum, pp 13--19

\bibitem[{Lin et~al(2016)Lin, Mausam, and Weld}]{lin2016re}
Lin C, Mausam M, Weld D (2016) Re-active learning: Active learning with
  relabeling. In: Proceedings of the AAAI Conference on Artificial Intelligence

\bibitem[{Lin et~al(2023)Lin, Chen, Song, Liu, Li, and Jiang}]{lin2023tree}
Lin X, Chen X, Song L, et~al (2023) Tree based progressive regression model for
  watch-time prediction in short-video recommendation. arXiv preprint
  arXiv:230603392

\bibitem[{Little and Rubin(2019)}]{little2019statistical}
Little RJ, Rubin DB (2019) Statistical analysis with missing data, vol 793.
  John Wiley \& Sons

\bibitem[{Ma et~al(2018)Ma, Zhao, Huang, Wang, Hu, Zhu, and Gai}]{ma2018entire}
Ma X, Zhao L, Huang G, et~al (2018) Entire space multi-task model: An effective
  approach for estimating post-click conversion rate. In: Proceedings of the
  International ACM SIGIR Conference on Research \& Development in Information
  Retrieval, pp 1137--1140

\bibitem[{McCallum et~al(1998)McCallum, Nigam et~al}]{mccallum1998employing}
McCallum A, Nigam K, et~al (1998) Employing {EM} and pool-based active learning
  for text classification. In: ICML, pp 350--358

\bibitem[{Mohan et~al(2013)Mohan, Pearl, and Tian}]{mohan2013graphical}
Mohan K, Pearl J, Tian J (2013) Graphical models for inference with missing
  data

\bibitem[{Nguyen et~al(2022)Nguyen, Ho, Xu, Dinh, and Nguyen}]{NGUYEN2022242}
Nguyen CV, Ho LST, Xu H, et~al (2022) Bayesian active learning with abstention
  feedbacks. Neurocomputing 471:242--250

\bibitem[{Nguyen et~al(2020)Nguyen, Shi, Ramakrishnan, Weinsberg, Lin, Metz,
  Chandra, Jing, and Kalimeris}]{nguyen2020clara}
Nguyen VA, Shi P, Ramakrishnan J, et~al (2020) {CLARA}: confidence of labels
  and raters. In: Proceedings of the 26th ACM SIGKDD International Conference
  on Knowledge Discovery \& Data Mining, pp 2542--2552

\bibitem[{Rosales et~al(2012)Rosales, Cheng, and Manavoglu}]{rosales2012post}
Rosales R, Cheng H, Manavoglu E (2012) Post-click conversion modeling and
  analysis for non-guaranteed delivery display advertising. In: Proceedings of
  the fifth ACM international conference on Web search and data mining, pp
  293--302

\bibitem[{Rubin(1976)}]{rubin1976inference}
Rubin DB (1976) Inference and missing data. Biometrika 63(3):581--592

\bibitem[{Settles(2009)}]{settles2009active}
Settles B (2009) Active learning literature survey technical report. University
  of Wisconsin-Madison Department of Computer Sciences

\bibitem[{Settles(2012)}]{settles2012uncertainty}
Settles B (2012) Uncertainty sampling. In: Active Learning. Springer, p 11--20

\bibitem[{Seung et~al(1992)Seung, Opper, and Sompolinsky}]{seung1992query}
Seung HS, Opper M, Sompolinsky H (1992) Query by committee. In: Proceedings of
  the fifth annual workshop on Computational learning theory, pp 287--294

\bibitem[{Sheng et~al(2008)Sheng, Provost, and Ipeirotis}]{sheng2008get}
Sheng VS, Provost F, Ipeirotis PG (2008) Get another label? improving data
  quality and data mining using multiple, noisy labelers. In: Proceedings of
  the 14th ACM SIGKDD international conference on knowledge discovery and data
  mining, pp 614--622

\bibitem[{Stekhoven and Bühlmann(2012)}]{stekhoven2012missforest}
Stekhoven DJ, Bühlmann P (2012) Missforest—non-parametric missing value
  imputation for mixed-type data. Bioinformatics 28(1):112--118

\bibitem[{Tax et~al(2021)Tax, de~Vries, de~Jong, Dosoula, van~den Akker, Smith,
  Thuong, and Bernardi}]{tax2021machine}
Tax N, de~Vries KJ, de~Jong M, et~al (2021) Machine learning for fraud
  detection in e-commerce: A research agenda. In: Deployable Machine Learning
  for Security Defense: Second International Workshop, MLHat 2021, Virtual
  Event, August 15, 2021, Springer, pp 30--54

\bibitem[{Tianchi(2018)}]{taobao}
Tianchi (2018) Ad display/click data on taobao.com.
  \urlprefix\url{https://tianchi.aliyun.com/dataset/dataDetail?dataId=56}

\bibitem[{Tifrea et~al(2023)Tifrea, Clarysse, and Yang}]{TifreaCY2}
Tifrea A, Clarysse J, Yang F (2023) Margin-based sampling in high dimensions:
  When being active is less efficient than staying passive. In: International
  Conference on Machine Learning (ICML), vol 202. {PMLR}, pp 34222--34262

\bibitem[{Yan et~al(2015)Yan, Chaudhuri, and Javidi}]{yan_2015}
Yan S, Chaudhuri K, Javidi T (2015) Active learning from noisy and abstention
  feedback. In: 53rd Annual Allerton Conference on Communication, Control, and
  Computing (Allerton), pp 1352--1357

\bibitem[{Yan et~al(2016)Yan, Chaudhuri, and Javidi}]{NIPS2016_dd77279f}
Yan S, Chaudhuri K, Javidi T (2016) Active learning from imperfect labelers.
  In: Advances in Neural Information Processing Systems

\bibitem[{Yang and Loog(2018)}]{yang2018benchmark}
Yang Y, Loog M (2018) A benchmark and comparison of active learning for
  logistic regression. Pattern Recognition 83:401--415

\bibitem[{Zhao et~al(2011)Zhao, Sukthankar, and
  Sukthankar}]{zhao2011incremental}
Zhao L, Sukthankar G, Sukthankar R (2011) Incremental relabeling for active
  learning with noisy crowdsourced annotations. In: 2011 IEEE third
  international conference on privacy, security, risk and trust and 2011 IEEE
  third international conference on social computing, IEEE, pp 728--733

\end{thebibliography}

\end{document}